\documentclass{article}
\PassOptionsToPackage{numbers, sort&compress}{natbib}
\usepackage[preprint]{dms}

\usepackage[utf8]{inputenc} 
\usepackage[T1]{fontenc}    
\usepackage{hyperref}       
\usepackage{url}            
\usepackage{booktabs}       
\usepackage{amsfonts}       
\usepackage{nicefrac}       
\usepackage{microtype}      
\usepackage{xcolor}         
\usepackage{subcaption}     
\usepackage{graphicx}       
\usepackage{siunitx}
\usepackage{float}
\usepackage{pifont}

\title{Procrustean Bed for AI-Driven Retrosynthesis: A Unified Framework for Reproducible Evaluation}
\author{%
  Anton Morgunov\thanks{Work performed in adherence with \href{https://github.com/ischemist/protocol}{isChemist Protocol}} \\
  Yale University \\
  \texttt{anton@ischemist.com} \\
  \And
  Victor S. Batista \\
  Yale University \\
  \texttt{victor.batista@yale.edu} \\
}

\begin{document}

\maketitle

\begin{abstract}
  Progress in computer-aided synthesis planning (CASP) is obscured by the lack of standardized evaluation infrastructure and the reliance on metrics that prioritize topological completion over chemical validity. We introduce \href{https://github.com/ischemist/project-procrustes}{RetroCast}, a unified evaluation suite that standardizes heterogeneous model outputs into a common schema to enable statistically rigorous, apples-to-apples comparison. The framework includes a reproducible benchmarking pipeline with stratified sampling and bootstrapped confidence intervals, accompanied by \href{https://syntharena.ischemist.com/}{SynthArena} (syntharena.ischemist.com), an interactive platform for qualitative route inspection. We utilize this infrastructure to evaluate leading search-based and sequence-based algorithms on a new suite of standardized benchmarks. Our analysis reveals a divergence between "solvability" (stock-termination rate) and route quality; high solvability scores often mask chemical invalidity or fail to correlate with the reproduction of experimental ground truths. Furthermore, we identify a "complexity cliff" in which search-based methods, despite high solvability rates, exhibit a sharp performance decay in reconstructing long-range synthetic plans compared to sequence-based approaches. We release the full framework, benchmark definitions, and a standardized database of model predictions to support transparent and reproducible development in the field
\end{abstract}

\section{Introduction}

We distinguish between two fundamental classes of scientific problems to which machine learning is applied: quantitative and structural. Quantitative problems, such as predicting drug toxicity, are defined by scalar targets and often constrained by data scarcity, analogous to early NLP challenges like sentiment analysis. In contrast, structural problems, like language modeling or protein folding, require generating complex objects governed by an underlying grammar. The most transformative successes of AI, from large language models~\cite{fewshot,r1,gpt4,agisparks} to AlphaFold~\cite{af2,af3,afimpact,afimpact2}, have occurred in these structural domains; foundation models trained on the structure of language, for example, now excel at sentiment analysis with little to no task-specific fine-tuning. We contend that mastery of structure is a prerequisite for solving downstream quantitative tasks. In organic chemistry, the paramount structural challenge is designing a valid synthetic pathway to a molecule of interest. This capability, retrosynthesis, is the key to unlocking critical quantitative problems like predicting a molecule's synthetic feasibility, a significant bottleneck in synthesis-aware virtual screening. Current accessibility heuristics, however, bypass the core structural challenge, relying on learned patterns that correlate with accessibility without ever generating the synthetic pathway itself. This, we argue, is a fundamental limitation: a model cannot judge the difficulty of a journey it cannot first articulate.

The dominant paradigm for computational retrosynthesis follows a two-part framework: a single-step model proposes disconnections, and a search algorithm explores the resulting pathway space~\cite{casp_1969}. While both components have seen rapid progress~\cite{coley_rank_2017,wlnet_coley_2017,fusionretro_2023,dai_gln_2019,ss_template_2024,mcts_2018,retrostar,coley_rl_2019,grasp_2022,egmcts_2023,retrograph}, the field's primary measure of success, traditionally called \textit{solvability}, creates a disconnect between reported performance and practical utility. A route is deemed "solved" if all its terminal nodes exist in a predefined commercial stock, but this is a purely topological check that provides no guarantee of chemical validity for the intermediate steps. This represents a methodological departure from the field's early best practices, which incorporated dedicated networks to filter infeasible reactions~\cite{mcts_2018}. The field now operates on an implicit and unevaluated assumption that single-step predictors have learned all complex rules of chemical feasibility. Consequently, high scores can be achieved for routes containing chemically nonsensical steps, rewarding any topological path regardless of its plausibility. To avoid the misleading implication of "solving" a chemical problem, we will henceforth refer to this metric by a more precise term: the \textit{Stock-Termination Rate (STR)}.

Attempts to address this validity gap with proxy metrics, such as forward-prediction confidence or round-trip accuracy~\cite{multistepttl_2023,retrogfn_2024,synllama_2025}, substitute direct validation with a reliance on auxiliary models, inheriting their biases without establishing a standardized measure of plausibility. Beyond these issues, the utility of STR is further undermined by a profound lack of standardization in its most critical component: the starting material stock. Our survey of prominent CASP tools reveals that the stock sets used for evaluation vary by over three orders of magnitude, from physically in-stock compounds to massive "make-on-demand" virtual libraries (Table S1). This thousand-fold disparity makes direct comparison of reported STR scores between models unreliable and can obscure the true signal of a model's chemical intelligence.

The PaRoutes benchmark~\cite{paroutes_2022}, the first large-scale dataset of experimental routes extracted from patent literature, sought to bridge this validity gap by measuring a model's ability to reproduce known syntheses. It exposed a stark disparity: models reporting over 97\% STR could find ground-truth routes with only 35–50\% top-10 accuracy. This motivated new architectures, such as the sequence-to-sequence DirectMultiStep model, which demonstrated an inverse performance profile: substantially improved route reproduction at the cost of a slightly lower STR~\cite{dms_2025}. However, interpreting PaRoutes as a universal gold standard is challenging because its reference routes are constructed from reactions reported within single patents. The endpoints of these patent-derived syntheses are not necessarily commercial precursors; to quantify this, we compared the n5 evaluation stock against the ASKCOS Buyables set, a collection of compounds available for purchase from eMolecules, Sigma-Aldrich, Mcule, LabNetwork, and ChemBridge~\cite{askcos_2025,higherlev,shee2025figshare}. Of the 13,306 unique leaf molecules, only 7,513 (56\%) are present in the Buyables set; as a result only 4,279 of the 10,000 reference routes in n5 have all their leaves within this realistic commercial stock. Optimizing for exact reproduction may therefore incentivize learning patent-specific patterns rather than the general principles of synthesis from commercial precursors.

More fundamentally, a focus on exact reproduction is inherently conservative, penalizing the discovery of novel yet plausible pathways. The field is therefore caught in a methodological bind: an STR metric that rewards novelty at the expense of plausibility, and a reproduction metric that ensures plausibility at the expense of novelty. This leaves no reliable way to distinguish a plausible new route from a chemically unfeasible artifact, creating a critical measurement gap where the most desirable outcome—--a novel, valid synthesis—--cannot be properly evaluated.

This methodological challenge is compounded by a practical infrastructure gap that blocks rigorous, large-scale comparison. Any attempt at a meta-analysis requires developing bespoke parsers for the fundamentally incompatible output formats of different CASP tools (Fig.~\ref{fig:babel_of_formats}). This heterogeneity is exacerbated by the high computational cost of generating predictions. This constraint is not hypothetical: a recent transformer-based model, for instance, was evaluated on only a 240-molecule subset of PaRoutes due to "high computing time"\cite{multistepttl_2023}. Faced with both unreliable metrics and these significant practical hurdles, the field's leading groups often bypass large-scale automated evaluation entirely, resorting instead to small-scale, costly human expert validation for their final assessments\cite{retrochimera_2025}. Collectively, these issues make direct model comparison impractical, if not impossible, leaving the field without a clear way to measure progress.

To resolve these issues, we introduce a unified framework that transforms ad-hoc evaluation into a systematic, scalable, and community-driven effort. First, we present \textit{RetroCast}, an open-source software package that provides both a universal translation layer for heterogeneous model outputs and an automated pipeline for performing statistically robust comparative analysis, all while ensuring auditable data provenance with cryptographic manifests. Second, using this framework, we perform the first rigorous, apples-to-apples comparison of the field's leading models on a suite of new, curated benchmarks designed to provide high diagnostic signal at a fraction of the computational cost. Our analysis exposes how high STR scores can mask chemically implausible predictions and reveals the divergent architectural signatures that emerge when models are evaluated with a more chemically meaningful, multi-ground-truth protocol. Finally, we release not only our tools and benchmarks but also the complete, standardized prediction database as a reusable community asset, accompanied by \textit{SynthArena}, an interactive web platform for qualitative inspection of route predictions. This provides the shared infrastructure needed to catalyze a shift in the field’s evaluation criteria: from merely checking for stock termination to the far more meaningful challenge of generating chemically plausible synthetic plans.

\section{Results}

\subsection{A Unified Framework for Reproducible Evaluation}
\label{sec:framework}
\begin{figure}[ht]
    \centering
    \includegraphics[width=0.95\textwidth]{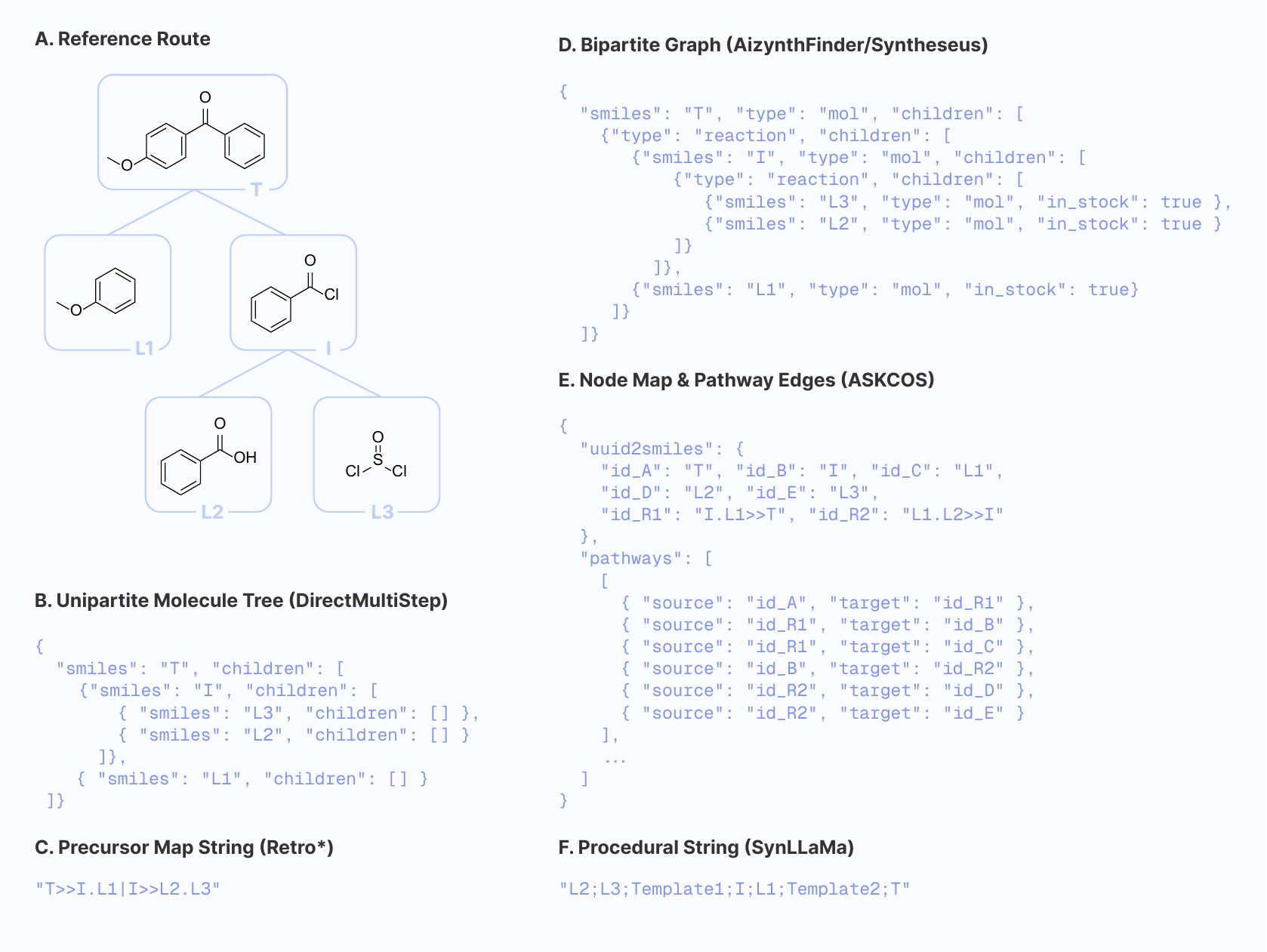}
    \caption{\textbf{The babel of retrosynthesis formats.} Illustration of five fundamentally incompatible output formats for a single reference synthetic route (A). Placeholders correspond to the target (T), an intermediate (I), and purchasable leaves (L1, L2, L3). Formats range from verbose, explicit graph structures (B, D, E) to concise, implicit string-based representations (C, F). (B) a simpler nested json of only molecule nodes, where reactions are implicit. (C) a declarative string mapping products to reactants. (D) a nested json where molecule nodes alternate with explicit reaction nodes. (E) a schema where a route is defined as a list of edges that reference a separate map of nodes. (F) a linear "recipe" string where the product of one step becomes an implicit reactant in the next. This heterogeneity necessitates a translation layer like RetroCast for any comparative analysis.}
    \label{fig:babel_of_formats}
\end{figure}

To resolve the field's infrastructure gap and enable rigorous comparison, we developed RetroCast, an open-source evaluation suite. Its foundation is a universal translation layer that addresses the heterogeneity of model outputs (Fig. \ref{fig:babel_of_formats}). Using an adapter-based architecture, RetroCast parses native formats into a single, standardized interchange schema, providing the necessary lingua franca for any cross-model analysis. We provide ready-to-use adapters for a comprehensive suite of tools, including search-based planners (AiZynthFinder~\cite{aizyn_2020}, Retro*~\cite{retrostar}, ASKCOS~\cite{askcos_2025}, Syntheseus~\cite{syntheseus_2024}, RetroChimera~\cite{retrochimera_2025}, DreamRetroEr~\cite{dreamretroer_2025}, MultiStepTTL~\cite{multistepttl_2023}, SynPlanner~\cite{synplanner_2024}), sequence-based models (DirectMultiStep~\cite{dms_2025}, SynLLaMa~\cite{synllama_2025}), and standard dataset formats like PaRoutes~\cite{paroutes_2022}. Beyond simple translation, the software provides a complete, automated pipeline for executing robust evaluations. Given a set of standardized predictions and a benchmark definition, RetroCast calculates key metrics, including Stock-Termination Rate and Top-K accuracy, with bootstrapped 95\% confidence intervals, and provides results stratified by route properties such as length and topology. To ensure all comparisons are auditable and reproducible, the entire workflow is accompanied by a system of cryptographic manifests; each stage of processing generates a manifest recording the SHA256 hashes of all inputs and outputs, creating a computationally verifiable chain of data provenance.

Building on this standardized data layer, we designed a principled evaluation protocol to address two distinct but often conflated goals: assessing a model's immediate practical utility versus enabling fair algorithmic comparison. To this end, we introduce two evaluation tracks, each with a corresponding suite of curated benchmarks derived from the PaRoutes n5 dataset (Table S2). The \textbf{Chemist-Aligned (\texttt{mkt-}) series} is designed to answer the practical question, "Which model is most useful today?" Models in this track have no training data restrictions and are evaluated against the ASKCOS Buyables stock, a realistic set of ~300k commercially available compounds. In contrast, the \textbf{Developer-Aligned (\texttt{ref-}) series} facilitates controlled comparison by asking, "Which algorithm is superior?" This track requires models to be trained on a common open-source dataset. To prevent data leakage, we propose that the entire PaRoutes n1 and n5 evaluation sets be excluded from any training corpus. Evaluation for this series uses the original patent-derived stocks from PaRoutes to isolate algorithmic performance from stock availability. All benchmarks were constructed using a stratified design based on route length and topology, providing high diagnostic signal at a fraction of the original's computational cost (details in Sec. \ref{sec:methods-bench}).

A critical ambiguity arises in the calculation of rank-dependent metrics like Top-K accuracy: which population of predictions should be ranked? The set of all raw model outputs, or only the subset that satisfies task-specific constraints? The choice can dramatically alter results and conclusions about model performance. Our framework resolves this by adopting a user-centric perspective: for a practicing chemist, a model's output is only useful if it represents a valid, coherent answer to the query. A route that fails to meet basic structural or user-specified constraints is not, by definition, a solution. This philosophy leads to a clear, sequential filtering protocol that must be applied before any ranking is performed: 1) all raw outputs are first filtered for structural integrity (e.g., valid SMILES, forming a directed acyclic graph); 2) the structurally sound pool is then filtered against all explicit task constraints, such as stock-termination; 3) finally, Top-K accuracy is calculated on this fully validated set, strictly preserving the model's original output order. This protocol ensures a model is judged on its ability to both produce and correctly rank valid solutions: the only relevant measure of performance from a user's perspective. This provides a stable and extensible foundation. As the field advances toward more complex queries, such as specifying starting materials~\cite{desp_2024,dms_2025} or forbidden reactions, our user-centric maxim offers a consistent path forward: first, filter for all constraints; then, and only then, evaluate the quality and ranking of the valid solutions.

To complement aggregate statistics with qualitative insight, our framework includes SynthArena, an open-source platform for interactive route inspection. SynthArena ingests standardized outputs from RetroCast and provides an interface for side-by-side route comparison, difference overlays, and annotation of commercial availability for all leaf nodes. This capacity for direct inspection allowed us to discover that some routes, marked incorrect by Top-K accuracy, are in fact shorter sub-routes of the reference pathway that terminated at commercially available intermediates. This valid outcome that was incorrectly penalized by a strict single ground-truth metric directly motivated the development of our more chemically meaningful multi-ground-truth evaluation protocol. To serve the community, we host a public instance at \href{https://syntharena.ischemist.com}{syntharena.ischemist.com} as a central "arena" and living leaderboard. As the platform is fully open-source, developers can also deploy it locally as a powerful tool for day-to-day model development, such as comparing checkpoints or diagnosing the effects of architectural changes. Our goal is to transform evaluation from a static, periodic exercise into a dynamic, ongoing process of collective error analysis and model improvement.

\subsection{Stock-Termination Rate is a Misleading Signal of Chemical Validity}
\label{sec:deconstructing-solvability}
Our unified analysis of prominent open-source models on the public USPTO-190 benchmark (evaluated with ASKCOS Buyables as a stock set) reveals that a high Stock-Termination Rate (STR) can be a misleading signal of chemical validity. While a conventional analysis would suggest a clear performance hierarchy, with the original Retro* model achieving a dominant 73.2\% STR (Table \ref{tab:pto-190}), systematic qualitative inspection of the underlying routes reveals a critical flaw in the metric. Because STR validates only the commercial availability of a route's terminal nodes, it provides no guarantee of chemical plausibility for the intermediate steps.

\begin{table}[htbp]
\centering
\small
\setlength{\tabcolsep}{4pt} 
\begin{tabular}{
  l
  S[table-format=2.1] @{\,[\,} S[table-format=2.1] @{,\,} S[table-format=2.1] @{\,]}
  @{\quad}
  S[table-format=2.1] @{\,[\,} S[table-format=1.1] @{,\,} S[table-format=2.1] @{\,]}
  @{\quad}
  S[table-format=2.1] @{\,[\,} S[table-format=1.1] @{,\,} S[table-format=2.1] @{\,]}
  @{\quad}
  c
}
\toprule
\textbf{Model} & \multicolumn{3}{c}{\textbf{STR (\%)}} & \multicolumn{3}{c}{\textbf{Top-1 Acc. (\%)}} & \multicolumn{3}{c}{\textbf{Top-10 Acc.
(\%)}} & \textbf{Time/Target (s.)} \\
\midrule
Retro* (High) & 73.2 & 66.8 & 79.5 & 10.0 & 5.8 & 14.2 & 10.0 & 5.8 & 14.2 & 35.3 \\
Retro* & 44.7 & 37.9 & 51.6 & 7.9 & 4.2 & 12.1 & 7.9 & 4.2 & 12.1 & 11.0 \\
AiZynF Retro* (High) & 36.8 & 30.0 & 43.7 & 0.5 & 0.0 & 1.6 & 2.1 & 0.5 & 4.2 & 133.0  \\
AiZynF MCTS (High) & 33.7 & 26.8 & 40.5 & 1.6 & 0.0 & 3.7 & 2.1 & 0.5 & 4.2 & 41.8 \\
AiZynF Retro* & 32.1 & 25.3 & 38.4 & 1.1 & 0.0 & 2.6 & 2.1 & 0.5 & 4.2 & 35.0 \\
Syntheseus LocalRetro & 33.7 & 26.8 & 40.5 & 0.0 & 0.0 & 0.0 & 0.0 & 0.0 & 0.0 & 18.4 \\
DMS Explorer XL & 29.5 & 23.2 & 35.8 & 0.5 & 0.0 & 1.6 & 1.1 & 0.0 & 2.6 & 20.5 \\
AiZynF MCTS & 24.7 & 18.4 & 31.1 & 1.1 & 0.0 & 2.6 & 2.1 & 0.5 & 4.2 & 11.04  \\
ASKCOS & 17.4 & 12.1 & 23.2 & 0.5 & 0.0 & 1.6 & 1.6 & 0.0 & 3.7 & 30.2 \\
\bottomrule
\end{tabular}
\vspace{0.3em}
\caption{\textbf{High stock-termination scores on the USPTO-190 benchmark often mask underlying challenges in chemical validity.} Performance of major retrosynthesis models on the 190-target USPTO test set, evaluated using the ASKCOS Buyables stock. Metrics include stock termination rate (STR, the fraction of targets for which a route to purchasable starting materials was found) and Top-K accuracy (the fraction of targets for which a reference route was found). \texttt{(High)} denotes a 500-iteration search vs. the 100-iteration default. The evaluated Retro* implementation returns a single route, making its Top-1 and Top-10 accuracy values identical. Values in brackets indicate bootstrapped 95\% confidence intervals. The full, interactive leaderboard for this benchmark is available on \href{https://syntharena.ischemist.com/leaderboard?benchmarkId=cmisbzsr30000xvdd613ymmbx}{SynthArena: link}. See Sec.~\ref{sec:methods-exec} for hardware specifications.}
\label{tab:pto-190}
\end{table}

\begin{figure}[ht]
  \centering
  \includegraphics[width=\textwidth]{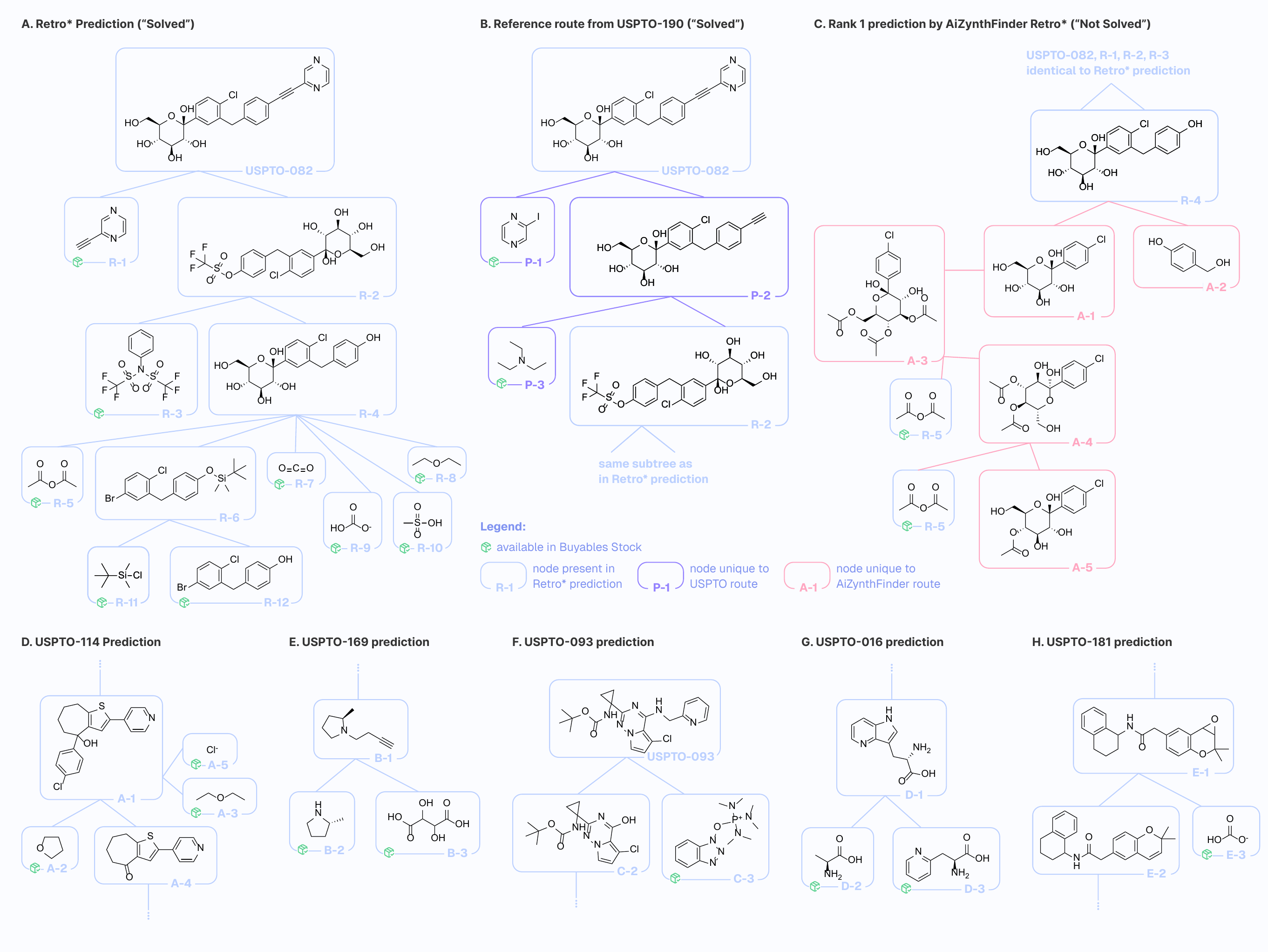}
  \caption{\textbf{High stock-termination rate rewards chemically invalid routes.} Analysis reveals how metrics blind to chemical validity can mislead. (\textbf{A-C}) The case of target USPTO-082. \textbf{A}, A "solved" route from the top-performing model hinges on a chemically implausible seven-reactant step. \textbf{B}, The official reference route contains the identical flawed transformation, showing the model's success is an artifact of pattern-matching corrupted data. \textbf{C}, A route from a newer model that avoids the nonsensical step is penalized for failing to find a "solved" path. (\textbf{D-H}) A catalog of chemical hallucinations from other "solved" routes, demonstrating the systemic nature of the issue. Violations include: \textbf{D}, mass balance error (un-sourced chloro-phenyl group); \textbf{E}, implausible transformation (tartaric acid as a propargyl source); \textbf{F}, mass balance error (un-sourced pyridylmethylamine); \textbf{G}, implausible reaction (amino acids to tryptophan core); \textbf{H}, unspecified reagent (epoxidation with carbonic acid). These cases show that naive stock termination rate fails to capture fundamental chemical principles. Interactive versions are on SynthArena: \href{https://syntharena.ischemist.com/benchmarks/cmisbzsr30000xvdd613ymmbx/targets/cmisbzt2900y4xvddbnu3q2k5?mode=pred-vs-pred&model1=cmise2ax00000qsddkfge5au3&rank1=1&model2=cmisdw7p10000ceddz6l01zhq&rank2=1}{USPTO-082}, \href{https://syntharena.ischemist.com/runs/cmise2ax00000qsddkfge5au3?stock=qhi67k3yqgqhrx49sc3akbih&target=cmisbzt5j01bwxvddy4a5xpu2&rank=1&search=114}{USPTO-114}, \href{https://syntharena.ischemist.com/runs/cmise2ax00000qsddkfge5au3?stock=qhi67k3yqgqhrx49sc3akbih&target=cmisbztag020hxvdd8nl7zg94&rank=1&search=169}{USPTO-169}, \href{https://syntharena.ischemist.com/runs/cmise2ax00000qsddkfge5au3?stock=qhi67k3yqgqhrx49sc3akbih&target=cmisbzt3h0139xvddt5rm50se&rank=1&search=93}{USPTO-93}, \href{https://syntharena.ischemist.com/runs/cmise2ax00000qsddkfge5au3?stock=qhi67k3yqgqhrx49sc3akbih&target=cmisbzsv10066xvddmu0bi5nk&rank=1&search=16}{USPTO-16}, \href{https://syntharena.ischemist.com/runs/cmise2ax00000qsddkfge5au3?stock=qhi67k3yqgqhrx49sc3akbih&target=cmisbztbj025gxvddwrx3reh6&rank=1&search=181}{USPTO-181}.}
  \label{fig:invalid-routes}
\end{figure}

This measurement gap is exemplified by the analysis of target USPTO-082 (Fig. \ref{fig:invalid-routes}A-C). The top-performing model's "solved" route is predicated on a chemically implausible seven-reactant combination. This is not a model hallucination but an exact reproduction of a flawed transformation present in the benchmark's own reference route. The model is thus rewarded for accurately pattern-matching corrupted data, while newer models that avoid this nonsensical step are penalized, failing to achieve a "solved" status.

The failure to penalize invalid chemistry is a systemic issue, not an isolated artifact. Fig. \ref{fig:invalid-routes}D-H presents a catalog of similarly unsound transformations extracted from five other "solved" routes generated by the same model. These examples document apparent violations of fundamental chemical principles, including mass balance errors and chemically nonsensical conversions. By incentivizing the discovery of any topological path to a commercial stock, STR systematically fails to capture a model's understanding of chemistry. Consequently, high STR scores alone---whether on this benchmark or any other---cannot be reliably interpreted as a signal of a model's ability to generate chemically sound synthetic plans.

\subsection{Multi-Ground-Truth Evaluation Reveals Divergent Architectural Signatures}
\label{sec:fallacy-of-gt}
The machine learning concept of a single "ground truth" is an oversimplification in organic synthesis, where multiple valid pathways to a target often exist. Current benchmarks, however, rigidly evaluate against a single patent-derived reference, incorrectly penalizing models for identifying shorter, more efficient syntheses that terminate at commercially available intermediates---a frequent occurrence that became apparent during qualitative inspection with SynthArena.

\begin{table}[htbp]
\centering
\small
\setlength{\tabcolsep}{4pt} 
\begin{tabular}{
  l
  S[table-format=3.1] @{\,[\,} S[table-format=3.1] @{,\,} S[table-format=3.1] @{\,]}
  @{\quad}
  S[table-format=3.1] @{\,[\,} S[table-format=2.1] @{,\,} S[table-format=2.1] @{\,]}
  @{\quad}
  S[table-format=3.1] @{\,[\,} S[table-format=2.1] @{,\,} S[table-format=2.1] @{\,]}
  @{\quad}
  c
}
\toprule
\textbf{Model} & \multicolumn{3}{c}{\textbf{STR (\%)}} & \multicolumn{3}{c}{\textbf{Top-1 Acc. (\%)}} & \multicolumn{3}{c}{\textbf{Top-10 Acc.
(\%)}} & \textbf{Time/Target (s.)} \\
\midrule
DMS Explorer XL & 96.3 & 93.1 & 98.8 & 33.8 & 26.9 & 41.3 & 57.5 & 50.0 & 65.0 & 17.9 \\
AiZynF MCTS & 98.1 & 95.6 & 100.0 & 21.3 & 15.0 & 27.5 & 41.3 & 33.8 & 48.8 & 6.7 \\
AiZynF MCTS (High) & 99.4 & 98.1 & 100.0 & 20.6 & 14.4 & 26.9 & 38.8 & 31.3 & 46.3 & 25.8 \\
Retro* (High) & 100.0 & 100.0 & 100.0 & 33.8 & 26.3 & 41.3 & 33.8 & 26.3 & 41.3 & 1.2 \\
Retro* & 98.8 & 96.9 & 100.0 & 33.1 & 25.6 & 40.6 & 33.1 & 25.6 & 40.6 & 0.8 \\
AiZynF Retro* & 98.8 & 96.9 & 100.0 & 8.1 & 4.4 & 12.5 & 27.5 & 20.6 & 34.4 & 32.7 \\
AiZynF Retro* (High) & 99.4 & 98.1 & 100.0 & 6.9 & 3.1 & 11.3 & 23.8 & 17.5 & 30.6 & 122.8 \\
Syntheseus LocalRetro & 95.6 & 92.5 & 98.8 & 8.8 & 4.4 & 13.1 & 20.6 & 14.4 & 26.9 & 15.8 \\
ASKCOS & 93.1 & 88.8 & 96.9 & 6.3 & 3.1 & 10.0 & 19.4 & 13.1 & 25.6 & 30.0 \\
\bottomrule
\end{tabular}
\vspace{0.3em}
\caption{\textbf{MGT evaluation reveals divergent architectural signatures on convergent routes.} Performance on the \texttt{mkt-cnv-160} benchmark, where the reference set is expanded to include all valid, commercially-terminated sub-routes. The results expose two distinct profiles: search-based models achieve near-perfect stock-termination rates (STR), while the sequence-based model shows substantially higher route-matching accuracy. The evaluated Retro* implementation returns a single route, making its Top-1 and Top-10 accuracy values identical. Brackets indicate 95\% confidence intervals. The full, interactive leaderboard is available on \href{https://syntharena.ischemist.com/leaderboard?benchmarkId=cmisc0flu0000boddjstwifeo}{SynthArena: link}}
\label{tab:mkt-cnv-160}
\end{table}

To create a more chemically meaningful benchmark, we expand the set of acceptable solutions. Our approach includes not only the full experimental sequence but also any of its constituent sub-routes that terminate exclusively in commercially available precursors (details Sec. \ref{sec:pruning}). This represents a pragmatic first step toward a more comprehensive evaluation. While it cannot yet reward entirely novel valid pathways, it provides a principled way to expand the reference set without sacrificing the chemical plausibility guaranteed by the original experiment. We refer to this as a Multi-Ground-Truth (MGT) evaluation, reflecting this expansion of the target set.

We applied this MGT protocol to our Chemist-Aligned benchmarks, which are evaluated against the ASKCOS Buyables stock. On the \texttt{mkt-cnv-160} benchmark of convergent routes, this reveals two distinct and often opposing performance profiles (Table \ref{tab:mkt-cnv-160}). Search-based models consistently achieve near-perfect stock-termination rates. This, however, is an intrinsic property of their design; the algorithm is engineered to explore until a stock set is reached, making high STR a satisfaction of a stopping condition rather than an independent measure of the route's quality. In contrast, the sequence-based DirectMultiStep model, for which stock enforcement is a post-processing step, exhibits the inverse profile: slightly lower STR but substantially higher accuracy in matching the structural patterns of known, plausible syntheses. This divergence underscores that stock termination and route reproduction are measuring fundamentally different capabilities.

This performance dichotomy is not an artifact of convergent topologies; an identical pattern is observed on the linear routes of the \texttt{mkt-lin-500} benchmark (Table \ref{tab:mkt-lin-500}). The results consistently show that models optimized for topological search and models that learn holistic route structure are being driven toward different solutions, a critical distinction obscured by previous evaluation paradigms.

\begin{table}[htbp]
\centering
\small
\setlength{\tabcolsep}{4pt} 
\begin{tabular}{
  l
  S[table-format=3.1] @{\,[\,} S[table-format=2.1] @{,\,} S[table-format=3.1] @{\,]}
  @{\quad}
  S[table-format=3.1] @{\,[\,} S[table-format=2.1] @{,\,} S[table-format=2.1] @{\,]}
  @{\quad}
  S[table-format=3.1] @{\,[\,} S[table-format=2.1] @{,\,} S[table-format=2.1] @{\,]}
  @{\quad}
  c
}
\toprule
\textbf{Model} & \multicolumn{3}{c}{\textbf{STR (\%)}} & \multicolumn{3}{c}{\textbf{Top-1 Acc. (\%)}} & \multicolumn{3}{c}{\textbf{Top-10 Acc.
(\%)}} & \textbf{Time/Target (s.)}\\
\midrule
DMS Explorer XL & 97.2 & 95.6 & 98.6 & 31.4 & 27.4 & 35.6 & 55.4 & 51.0 & 59.8 & 14.5 \\
AiZynF MCTS & 97.6 & 96.2 & 98.8 & 17.8 & 14.4 & 21.2 & 35.8 & 31.6 & 40.0 & 6.1 \\
AiZynF MCTS (High) & 98.4 & 97.2 & 99.4 & 17.4 & 14.2 & 20.8 & 33.4 & 29.2 & 37.6 & 20.8 \\
AiZynF Retro* & 98.0 & 96.6 & 99.2 & 10.4 & 7.8 & 13.2 & 28.6 & 24.6 & 32.6 & 26.3 \\
AiZynF Retro* (High) & 99.0 & 98.0 & 99.8 & 9.2 & 6.8 & 11.8 & 25.4 & 21.6 & 29.2 & 103.1 \\
Retro* (High) & 99.8 & 99.4 & 100.0 & 22.2 & 18.6 & 25.8 & 22.2 & 18.6 & 25.8 & 0.6 \\
Retro* & 99.8 & 99.4 & 100.0 & 22.2 & 18.6 & 25.8 & 22.2 & 18.6 & 25.8 & 0.5 \\
Syntheseus LocalRetro & 94.0 & 91.8 & 96.0 & 8.6 & 6.2 & 11.2 & 18.8 & 15.4 & 22.4 & 11.9 \\
ASKCOS & 94.4 & 92.4 & 96.4 & 6.2 & 4.2 & 8.4 & 16.2 & 13.2 & 19.6 & 29.5 \\
\bottomrule
\end{tabular}
\vspace{0.3em}
\caption{\textbf{The performance dichotomy persists on linear routes.} Performance on the \texttt{mkt-lin-500} benchmark using MGT evaluation. The results confirm the findings from the convergent set, with models optimized for topological search (high STR) and models that learn holistic route structure (high accuracy) exhibiting inverse performance profiles. This demonstrates the robustness of the observation. The evaluated Retro* implementation returns a single route, making its Top-1 and Top-10 accuracy values identical. Brackets indicate 95\% confidence intervals. The full, interactive leaderboard is available on \href{https://syntharena.ischemist.com/leaderboard?benchmarkId=cmisc0cnd0000a8dd4g4pdf0s}{SynthArena: link}}
\label{tab:mkt-lin-500}
\end{table}

\begin{table}[htbp]
\centering
\small
\setlength{\tabcolsep}{4pt} 
\begin{tabular}{
  l
  S[table-format=2.1] @{\,[\,} S[table-format=2.1] @{,\,} S[table-format=2.1] @{\,]}
  @{\quad}
  S[table-format=2.1] @{\,[\,} S[table-format=2.1] @{,\,} S[table-format=2.1] @{\,]}
  @{\quad}
  S[table-format=2.1] @{\,[\,} S[table-format=2.1] @{,\,} S[table-format=2.1] @{\,]}
  @{\quad}
  S[table-format=2.1] @{\,[\,} S[table-format=2.1] @{,\,} S[table-format=2.1] @{\,]}
}
\toprule
\textbf{Model} & \multicolumn{3}{c}{\textbf{Length 2}} & \multicolumn{3}{c}{\textbf{Length 3}} & \multicolumn{3}{c}{\textbf{Length 4}} & \multicolumn{3}{c}{\textbf{Length 5}} \\
\midrule
AiZynF MCTS & 75.0 & 62.5 & 87.5 & 50.0 & 35.0 & 65.0 & 25.0 & 12.5 & 40.0 & 15.0 & 5.0 & 27.5 \\
AiZynF MCTS (High) & 80.0 & 67.5 & 92.5 & 37.5 & 22.5 & 52.5 & 25.0 & 12.5 & 40.0 & 12.5 & 2.5 & 22.5 \\
AiZynF Retro* & 65.0 & 50.0 & 80.0 & 17.5 & 7.5 & 30.0 & 12.5 & 2.5 & 22.5 & 15.0 & 5.0 & 27.5 \\
AiZynF Retro* (High) & 50.0 & 35.0 & 65.0 & 20.0 & 7.5 & 32.5 & 12.5 & 2.5 & 25.0 & 12.5 & 2.5 & 22.5 \\
ASKCOS & 37.5 & 22.5 & 52.5 & 20.0 & 7.5 & 32.5 & 5.0 & 0.0 & 12.5 & 15.0 & 5.0 & 27.5 \\
DMS Explorer XL & 75.0 & 60.0 & 87.5 & 57.5 & 42.5 & 72.5 & 45.0 & 30.0 & 60.0 & 52.5 & 37.5 & 67.5 \\
Retro* & 50.0 & 35.0 & 65.0 & 37.5 & 22.5 & 52.5 & 25.0 & 12.5 & 40.0 & 20.0 & 7.5 & 32.5 \\
Retro* (High) & 50.0 & 35.0 & 65.0 & 37.5 & 22.5 & 52.5 & 27.5 & 15.0 & 42.5 & 20.0 & 7.5 & 32.5 \\
Syntheseus LocalRetro & 40.0 & 25.0 & 55.0 & 22.5 & 10.0 & 35.0 & 10.0 & 2.5 & 20.0 & 10.0 & 2.5 & 20.0 \\
\bottomrule
\end{tabular}
\vspace{0.3em}
\caption{\textbf{Accuracy decays with complexity, exposing a measurement crisis.} Top-10 route-matching accuracy on the \texttt{mkt-cnv-160} benchmark, stratified by reference route length. Search-based models excel on short routes but their performance collapses as complexity increases, while the sequence-based model (DMS) remains more robust. This sharp decay reveals the fundamental limit of reference-based evaluation: it penalizes the discovery of novel routes as failures, making it impossible to distinguish a failed search from a creative success. The evaluated Retro* implementation returns a single route, making its Top-1 and Top-10 accuracy values identical. Brackets indicate 95\% confidence intervals. Detailed statistics are available for interactive exploration on \href{https://syntharena.ischemist.com/leaderboard?benchmarkId=cmisc0flu0000boddjstwifeo}{SynthArena: link}}

\label{tab:mkt-cnv-160-stratified}
\end{table}

\subsection{Stratified Analysis Uncovers a "Complexity Cliff"}
Stratifying the analysis by route length exposes the architectural signatures that aggregate metrics obscure. On the mkt-cnv-160 benchmark, search-based models excel at matching short reference routes, but their accuracy decays sharply as synthetic complexity increases (Table \ref{tab:mkt-cnv-160-stratified}). In contrast, the sequence-based DirectMultiStep model maintains more consistent performance on longer routes. This trend culminates in a "complexity cliff" on the \texttt{ref-lng-84} benchmark, a stress test of exclusively long routes (lengths 8-10), where the route-matching accuracy of all evaluated search-based models collapses to near-zero (Table S7).

This result presents two non-exclusive interpretations. The first is algorithmic: iterative tree search may be inherently ill-suited for the combinatorial complexity of long-range planning. The second, however, is a measurement limitation: a search algorithm succeeding in finding a novel, plausible route is penalized as a failure by any reference-based metric. This ambiguity exposes the ultimate limitation of the current evaluation paradigm. It is fundamentally incapable of distinguishing a failed search from a creative success, creating a critical measurement gap where the field's most important goal---the discovery of novel, valid syntheses---cannot be rewarded.

\subsection{The Cost-Performance Frontier in Synthesis Planning}
\begin{figure}[htbp]
  \centering
  \includegraphics[width=\textwidth]{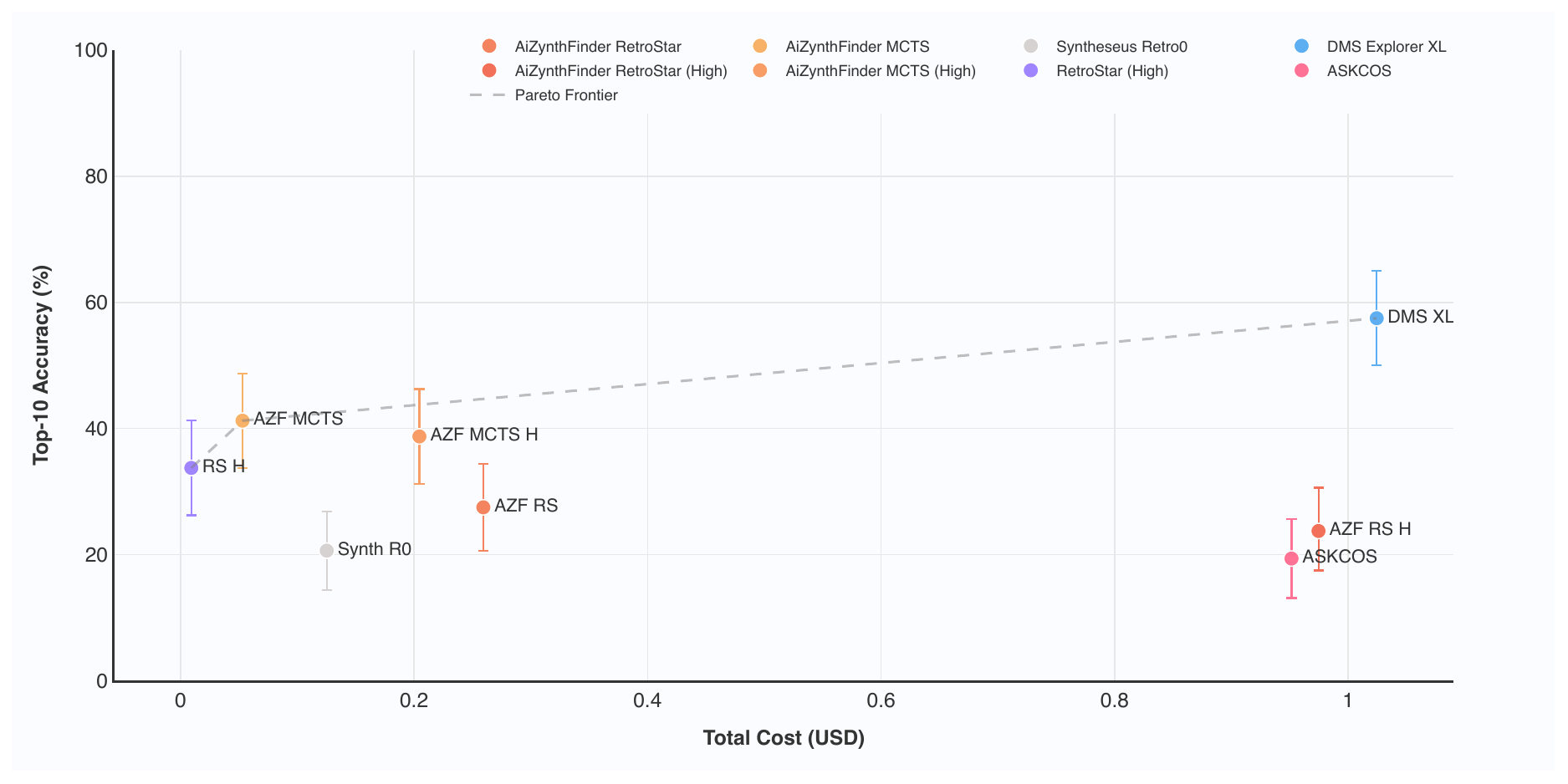}
  \caption{\textbf{The Economic Trade-offs of Synthesis Planning.} Pareto plot of Top-10 route-matching accuracy versus computational cost (USD) on the \texttt{mkt-cnv-160} benchmark. The efficient frontier (dashed line) illustrates the apparent optimal trade-off, but this landscape is defined by the field's measurement limitations: models in the low-cost region can be predicated on chemically implausible steps, while the accuracy metric itself penalizes the discovery of valid novel routes. Error bars represent 95\% bootstrapped CIs.}
  \label{fig:mkt-cnv-160-pareto}
\end{figure}

To provide a practical dimension to our analysis, we augment accuracy with computational cost, measured in USD based on cloud compute pricing (details in Sec. \ref{sec:methods-exec}). The resulting Pareto plot for the mkt-cnv-160 benchmark establishes a cost-performance frontier, quantifying the trade-off between predictive accuracy and economic cost (Fig. \ref{fig:mkt-cnv-160-pareto}).

This frontier, however, must be interpreted through the lens of our prior findings. First, the low-cost, high-STR region is occupied by models whose performance can be predicated on chemically implausible steps. Second, the accuracy metric itself penalizes novel discovery, potentially misclassifying exploratory models as inefficient when they are merely operating outside the benchmark's conservative definition of success.

These qualifications notwithstanding, this analysis establishes computational cost as a critical and quantifiable axis for evaluation. Existing models already show order-of-magnitude differences in cost, which at the scale of a synthesis-aware virtual screening campaign is the difference between a feasible project and an impossible one. As this cost-performance landscape is robust across route topologies (Fig. S1), it provides a necessary, if incomplete, framework for assessing the practical utility of synthesis planning tools.

\section{Discussion}
The evidence presented in this work demonstrates that the stock-termination rate is an incomplete and potentially misleading proxy for progress in retrosynthesis. Our analysis revealed that because the metric ignores the chemical validity of intermediate steps, it can reward the generation of chemically implausible routes. For the dominant class of search-based models, high STR scores are not an independent measure of a route's quality but rather a reflection of the algorithm's success in satisfying its core search objective. An over-reliance on such a metric may inadvertently steer the field's focus toward topological pathfinding at the expense of chemical plausibility. We therefore suggest the community reconsider its role, shifting its use from a primary performance objective to that of a necessary but insufficient post-hoc filter.

While route reproduction accuracy is a clear improvement over naive stock termination, we must acknowledge its primary limitation: it is inherently conservative. By defining correctness as adherence to a known experimental path, Top-K accuracy provides a crucial proxy for chemical plausibility but cannot, by definition, reward the discovery of a novel and potentially more efficient synthetic route. This presents a critical measurement challenge. As our stratified analysis showed, the sharp decline in Top-K accuracy for search-based models on complex targets could represent either a failure to find the known path or a success in discovering a valid alternative that our benchmark cannot recognize.

Disambiguating these outcomes—distinguishing a failed prediction from a creative one—is perhaps the central challenge for the next generation of retrosynthesis evaluation. Our public release of the complete, standardized prediction database is intended to catalyze this effort. By creating a reusable community asset, we decouple the development of novel plausibility metrics from the significant computational cost of running planning algorithms, allowing researchers to rapidly prototype and validate new scoring functions on a comprehensive set of state-of-the-art predictions. Furthermore, our SynthArena platform can be extended beyond a simple viewer into a system for distributed, expert annotation. By enabling chemists to flag and categorize invalid reaction steps, we can transform our static data release into a living, community-curated dataset of "chemical bugs," moving beyond passive metrics to an active, adversarial process of chemical bug hunting.

This focus on developing better plausibility metrics points toward a more rigorous standard for the field. Our work also demonstrates that large-scale benchmarks are not always necessary; smaller, carefully stratified benchmarks can provide greater diagnostic power by revealing performance boundaries invisible in a single, top-line number. To ensure these improved practices lead to sustainable progress, we advocate for the formal recognition of two distinct research tracks: one for methods and another for evaluation. This separation of concerns would discourage the practice of proposing new metrics alongside the methods they are designed to evaluate and ensure that progress is always measured against a stable, community-vetted yardstick.

Establishing a reliable measure of chemical plausibility is the key that unlocks a more sophisticated, multi-faceted evaluation of synthesis planning. The utility of secondary metrics, such as route diversity or average length, is currently limited; these measures are meaningless when calculated over a pool of predictions that includes chemically invalid pathways. Establishing a baseline of chemical validity, however, would transform these currently noisy measures into high-signal diagnostics. Only then can the field begin to ask more nuanced questions about the creativity, efficiency, and elegance of the chemically sound plans that different models produce.

\section{Methods}
\subsection{Benchmark and Stock Set Curation}
\label{sec:methods-bench}
Our evaluation framework is built upon a combination of curated benchmark subsets and precisely defined starting material stocks. For the Chemist-Aligned (\texttt{mkt-}) series, designed to assess practical utility, we standardized on the ASKCOS "Buyables" set~\cite{higherlev,shee2025figshare}. This stock contains 313 458 compounds available from eMolecules, Sigma-Aldrich, LabNetwork, Mcule, and ChemBridge, representing a realistic set of commercially acquirable starting materials. For the Developer-Aligned (\texttt{ref-}) series, designed for fair algorithmic comparison, we used the original stock sets from the PaRoutes publication, defined as the set of all unique leaf nodes from the corresponding evaluation set (n1 or n5). This follows the original protocol to isolate algorithmic performance from variations in stock availability.

All curated benchmarks were derived from the 10 000 experimental routes in the PaRoutes n5 evaluation set, except for \texttt{ref-lng-84} which used routes from n1 set as well. To construct the \texttt{mkt-} benchmarks, the full n5 set was first filtered to retain only routes where all leaf nodes are present in the ASKCOS Buyables stock, yielding a commercially relevant subset of 4 279 routes. The \texttt{ref-} benchmarks were derived from the full, unfiltered n5 set.

From these source sets, benchmarks were created using stratified random sampling based on route length and topology (linear vs. convergent). For instance, the \texttt{mkt-lin-500} benchmark was constructed by randomly sampling 100 routes from each length category (2, 3, 4, 5, and 6) from the commercially relevant linear route subset. This stratified design ensures that performance evaluation is not skewed by the natural prevalence of shorter routes in the source data and enables a direct, unweighted analysis of how model performance varies with synthetic complexity. To incorporate our multi-ground-truth evaluation paradigm while ensuring absolute reproducibility, the route pruning and expansion algorithm (described in Section \ref{sec:pruning}) was executed as a pre-computation step. The complete, expanded set of all valid ground-truth routes for each target is stored directly within the final benchmark definition file. This transforms the benchmark into a static, verifiable artifact and ensures that all subsequent evaluations are performed against an identical set of solutions

Recognizing that a single random sample could yield unrepresentative results, we implemented a rigorous seed selection protocol to ensure the statistical stability of our benchmarks. For each proposed benchmark, we generated 15 candidate subsets using 15 distinct random seeds. We then evaluated a reference model (DMS Explorer XL) on each of the 15 subsets to obtain performance metrics (Solvability, Top-1, and Top-10 accuracy). A deviation score was calculated for each seed, defined as the sum of the squared Z-scores for the three metrics relative to the mean performance across all 15 seeds. This score quantifies how much a given seed's resulting benchmark deviates from the average behavior. The seed with the lowest deviation score was selected to generate the final, canonical benchmark used in this study, providing strong evidence that our results are robust to sampling variance (See Fig. S2-S4). Finally, the USPTO-190 benchmark was used as published, with evaluation performed against the ASKCOS Buyables stock to maintain consistency with our mkt- series analysis.

\subsection{Statistical Analysis and Metrics}

\subsubsection{Confidence Intervals and Significance Testing}
\label{sec:methods-ci}
All reported metrics are accompanied by 95\% confidence intervals (CIs) calculated using a non-parametric bootstrap procedure with 10,000 resamples. For a given metric and a set of N targets, we generated 10,000 bootstrap samples by resampling the N target outcomes with replacement. The reported CI represents the 2.5th and 97.5th percentiles of the distribution of the means of these bootstrap samples. To guard against misleading CIs from small or skewed samples, we implemented a reliability check flagging estimates derived from sample sizes $N<30$ or where the number of positive or negative outcomes was less than 5, in line with standard statistical practice for proportions.

To determine if the performance difference between two models is statistically significant, we employed a paired bootstrap difference test. For a set of targets evaluated by both models, we first created a vector of the paired differences in their outcomes (e.g., 1 if model B succeeded and A failed, -1 if A succeeded and B failed, 0 otherwise). This vector of differences was then bootstrapped 10,000 times to construct a 95\% CI for the mean difference. A difference was considered statistically significant if the resulting 95\% CI did not contain zero.

\subsubsection{Refined Top-K Accuracy (Multi-Ground-Truth Evaluation)}
\label{sec:pruning}
The standard Top-K accuracy metric is overly rigid, incorrectly penalizing models for identifying valid, economically superior routes that terminate early at commercially available intermediates. To address this limitation, we developed a more chemically meaningful multi-ground-truth (MGT) evaluation protocol. For a given reference route from a benchmark and a specified commercial stock set, our algorithm generates an expanded set of valid ground-truth pathways through the following procedure:

\begin{enumerate}
    \item \textbf{Identify Pruning Points:} All intermediate molecules within the reference route that are also present in the commercial stock are identified as potential pruning points.
    
    \item \textbf{Generate Valid Sub-routes:} An expanded set of acceptable ground-truth routes is generated by treating these intermediates as alternative starting materials. To avoid generating redundant or invalid sub-routes (e.g., pruning at both an intermediate and its own precursor), the generation is constrained to combinations of intermediates that form an \textit{antichain} within the route's directed acyclic graph. An antichain is a set of nodes where no node is an ancestor of another, ensuring that each pruning point is independent.
    
    \item \textbf{Validate Stock Termination:} Each newly generated "pruned" route is validated to ensure all of its terminal leaf nodes are present in the commercial stock.
    
    \item \textbf{Evaluate Match:} A model's prediction is scored as a Top-K success if it achieves an exact topological match with \textit{any} route in this expanded set of valid ground truths (i.e., the original reference route plus all valid, solvable pruned variants).
\end{enumerate}

\subsection{Model Selection and Execution}
\label{sec:methods-exec}
The models evaluated in this study comprise all major open-source, self-deployable retrosynthesis planners available at the time of writing: AiZynthFinder, ASKCOS, Retro*, Syntheseus, SynPlanner, and DirectMultiStep. All model execution was automated via scripts provided in the RetroCast repository. To ensure full computational reproducibility, the environment was managed by the \texttt{uv} package manager, with a committed \texttt{uv.lock} file guaranteeing byte-for-byte identical versions of all dependencies.

To establish a baseline reflecting a standard installation, model configurations were set to their published defaults; all configuration files are committed to the repository for complete auditability. The sole deviation from default settings was an increase in the number of search iterations to 500 for all applicable models, aligning our protocol with the high-effort configuration used in the original PaRoutes study~\cite{paroutes_2022}. Execution was performed on cloud compute resources. Search-based planners were run on AWS EC2 \texttt{c7i.xlarge} instances (4 vCPUs, 8 GB RAM, \$0.1785/hr), with the more demanding ASKCOS planner run on a \texttt{c7i.4xlarge} instance (16 vCPU, 32 GB RAM, \$0.714/hr). The sequence-based DirectMultiStep model was run on NVIDIA A100 GPUs (40 GB SXM4, \$1.29/hr) provided by Lambda, Inc.

\section*{Code and Data Availability}

All code, configuration files, and analysis scripts are organized in the `RetroCast` Python package, which is open-source and publicly available on GitHub at \href{https://github.com/ischemist/project-procrustes}{github.com/ischemist/project-procrustes} under an MIT license.

The complete dataset generated and analyzed in this study—including all raw model outputs, standardized route data, and aggregated statistical results---is permanently archived and publicly accessible at \href{https://files.ischemist.com/retrocast/publication-data/}{files.ischemist.com/retrocast/publication-data/}. The integrity of the entire data archive can be computationally verified by executing the \verb|retrocast verify --all| command.

The interactive web platform, SynthArena, is also open-source, with its code available at \href{https://github.com/ischemist/syntharena}{github.com/ischemist/syntharena}. A public instance of the platform, hosting the results presented in this paper, is accessible at \href{https://syntharena.ischemist.com}{syntharena.ischemist.com}.

\section*{Author Contributions}

\noindent\textbf{Anton Morgunov:} Conceptualization, Methodology, Software, Validation, Formal Analysis, Investigation, Data Curation, Writing – Original Draft, Visualization. \\
\noindent\textbf{Victor S. Batista:} Supervision, Funding Acquisition, Writing – Review \& Editing, Project Administration.

\section*{Conflict of Interest}
A.M. is the primary author of the DirectMultiStep (DMS) model, one of the methods evaluated in this study. To ensure objectivity, the evaluation framework and all associated data are fully open-source, and all results are computationally reproducible via the provided scripts, allowing for independent verification of the findings. V.S.B. declares no competing interests.

\section*{Acknowledgement}

\noindent A.M. thanks Dr. Bogdan Zagribelnyy for insightful discussions. This research was supported by the National Science Foundation under Grant No. CHE-2124511. This research was also supported in part by Lambda, Inc.

\medskip
{
\small
\bibliography{references}
}

\newpage
\renewcommand{\thetable}{S\arabic{table}} 
\renewcommand{\thefigure}{S\arabic{figure}} 
\renewcommand{\thesection}{S\arabic{section}}
\setcounter{table}{0} 
\setcounter{figure}{0} 
\setcounter{section}{0} 

\section*{Supplementary Notes}

This document provides supplementary information to the main text. It includes detailed results tables, figures, and the full methods section.

\paragraph{Supplementary Tables S1 and S2}
Table \ref{tab:stock_sets}, referenced in the main text Introduction, details the wide variance in starting material stock sets used across prominent retrosynthesis models. Table \ref{tab:benchmarks} provides a summary of the curated benchmarks introduced in this work, including their size, description, and the stock set used for evaluation.

\paragraph{Supplementary Tables S3 and S4}
Tables \ref{tab:mkt-cnv-160-single-gt} and \ref{tab:mkt-lin-500-single-gt} provide the baseline evaluation results on the \texttt{mkt-cnv-160} and \texttt{mkt-lin-500} benchmarks, respectively, using a rigid single-ground-truth (SGT) paradigm. As discussed in the main text, these results systematically underestimate model performance by penalizing the discovery of valid, shorter sub-routes. They serve as the control against which the improved multi-ground-truth (MGT) evaluation (main text Tables 2 and 3) is compared.

\paragraph{Supplementary Tables S5, S6, and S7}
These tables present the full results for the Developer-Aligned (\texttt{ref-}) series of benchmarks. These benchmarks use the original patent-derived stocks from the PaRoutes dataset to facilitate fair algorithmic comparison, isolating model performance from the choice of commercial stock. Table \ref{tab:ref-lin-600} (\texttt{ref-lin-600}) and Table \ref{tab:ref-cnv-400} (\texttt{ref-cnv-400}) corroborate the findings from the Chemist-Aligned series. Table \ref{tab:ref-lng-84} (\texttt{ref-lng-84}) presents the results of the "complexity cliff" stress test on exclusively long routes, as discussed in the main text.

\paragraph{Caveat on Training Data Standardization:} The primary objective of the Developer-Aligned (\texttt{ref-}) series is to isolate algorithmic performance by controlling for the training corpus. However, to provide an immediate operational baseline, the results in Tables S5--S7 utilize the official pre-trained weights for each model. Since these models were originally trained on varying datasets, performance differences may currently reflect data discrepancies as well as architectural ones. We present these results as a provisional snapshot; we expect that as the community adopts this framework and populates the leaderboard with models retrained on the standardized split, these benchmarks will evolve into a pure measure of algorithmic superiority.

\begin{figure}[h]
  \centering
  \includegraphics[width=\textwidth]{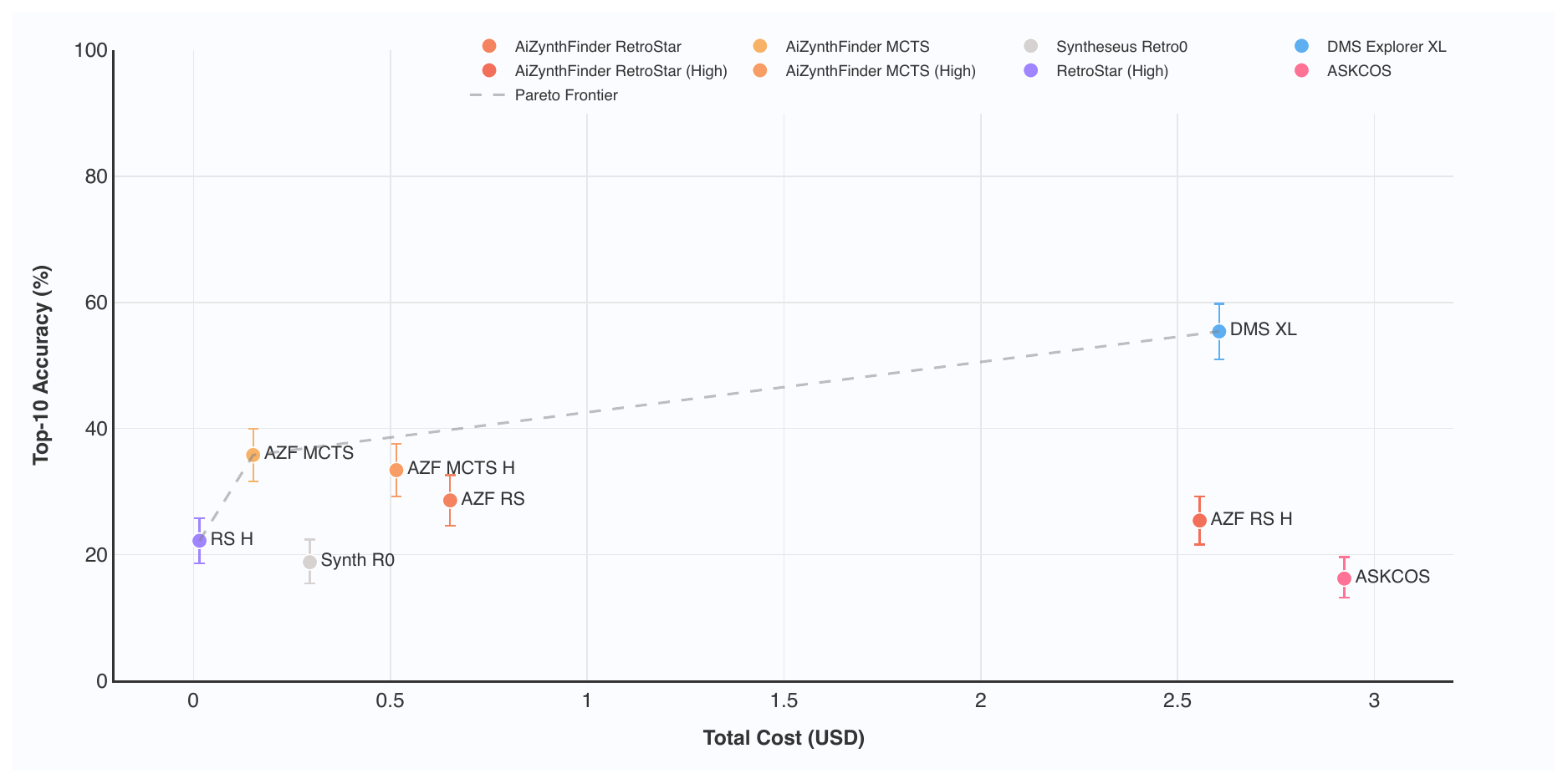}
  \caption{\textbf{The Cost of Accuracy on Linear Routes.} A Pareto plot of Top-10 Accuracy versus Total Cost (USD) on the \texttt{mkt-lin-500} benchmark. The analysis confirms the trade-off structure seen in Figure 3 is generalizable to linear routes. While absolute costs differ due to the larger benchmark size, the relative cost-performance profiles and the shape of the efficient frontier are consistent, demonstrating a robust relationship between accuracy and computational cost. Error bars are 95\% bootstrapped CIs.}
  \label{fig:mkt-lin-500-pareto}
\end{figure}

\paragraph{Supplementary Figure S1}
Figure~\ref{fig:mkt-lin-500-pareto} shows the Pareto plot of accuracy versus cost for the \texttt{mkt-lin-500} benchmark, demonstrating that the cost-performance trade-offs identified in the main text are robust across different route topologies.

\begin{figure}[h]
\centering
\includegraphics[width=\textwidth]{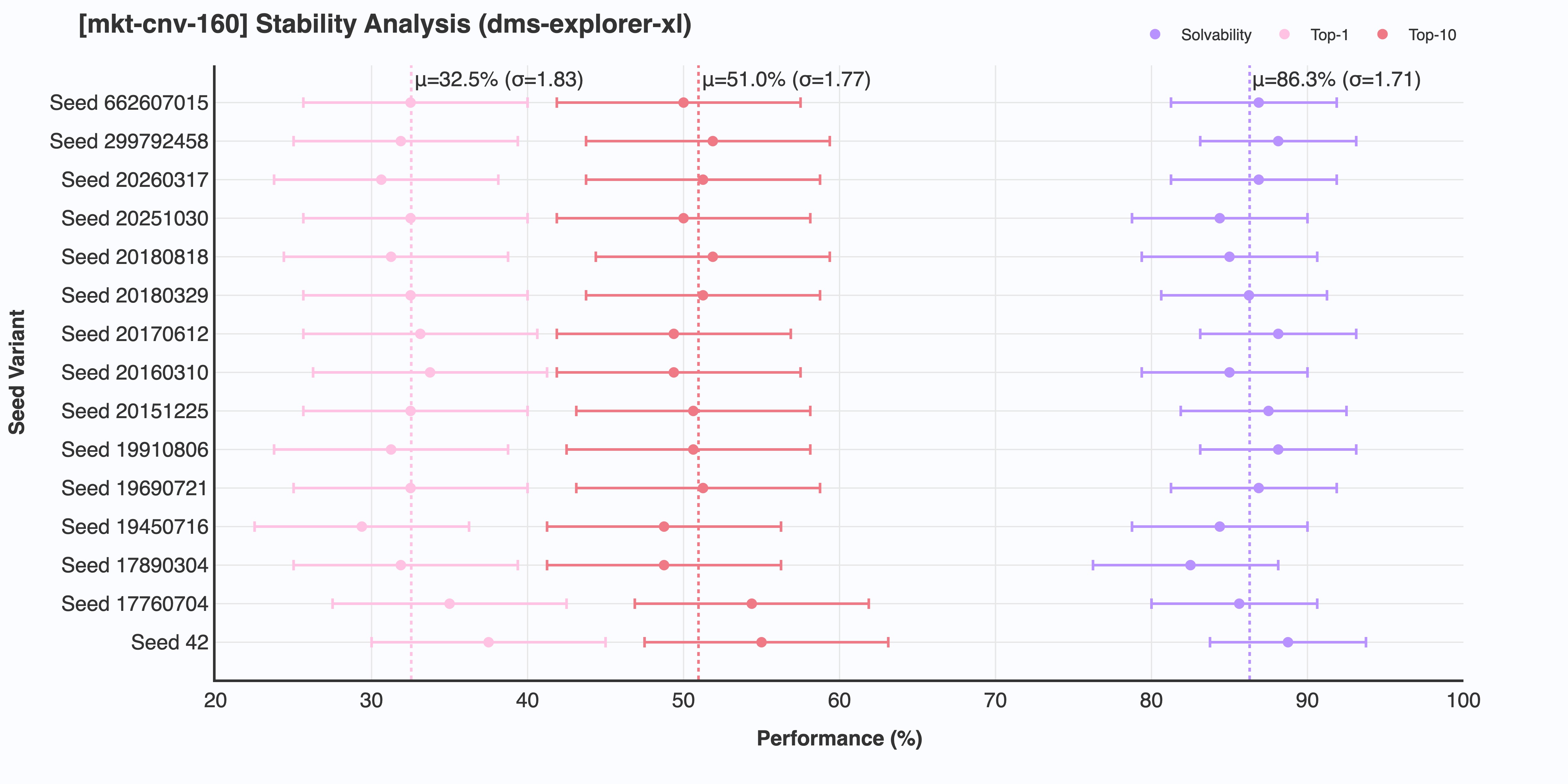}
\caption{\textbf{Selection of a statistically representative benchmark seed for \texttt{mkt-cnv-160}.} The plot shows the performance variation of a reference model (DMS Explorer XL) across 15 candidate benchmarks, each generated with a different random seed. Points indicate the mean accuracy (Solvability, Top-1, and Top-10), with horizontal bars representing the bootstrapped 95\% confidence intervals. Dashed vertical lines mark the grand mean performance across all seeds. This stability analysis allows us to quantify the variance introduced by subset sampling and select a seed (in this case, 20180329) that yields a benchmark whose metrics are demonstrably close to the central tendency, ensuring our evaluations are robust against sampling artifacts.}
\label{fig:mkt-cnv-160-seed-stability}
\end{figure}

\begin{figure}[h]
\centering
\includegraphics[width=\textwidth]{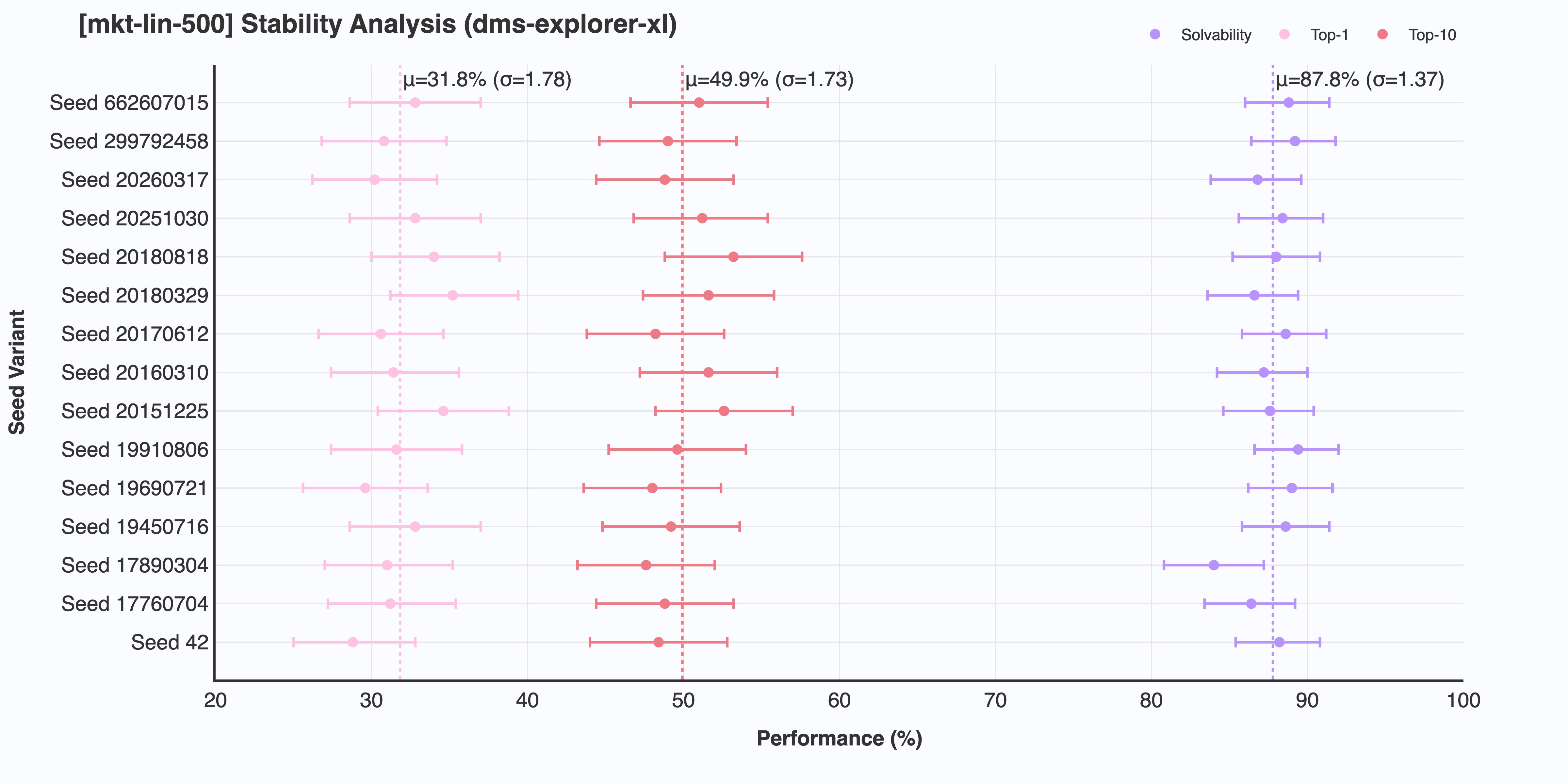}
\caption{\textbf{Benchmark stability analysis for \texttt{mkt-lin-500}.} Performance of a reference model is shown for 15 candidate benchmarks generated from different random seeds. The analysis confirms that while sampling introduces variance, a representative seed (19450716) can be selected by minimizing the deviation from the grand mean (dashed lines), thereby ensuring the benchmark is not an outlier.}
\label{fig:mkt-lin-500-seed-stability}
\end{figure}

\begin{figure}[h]
\centering
\includegraphics[width=\textwidth]{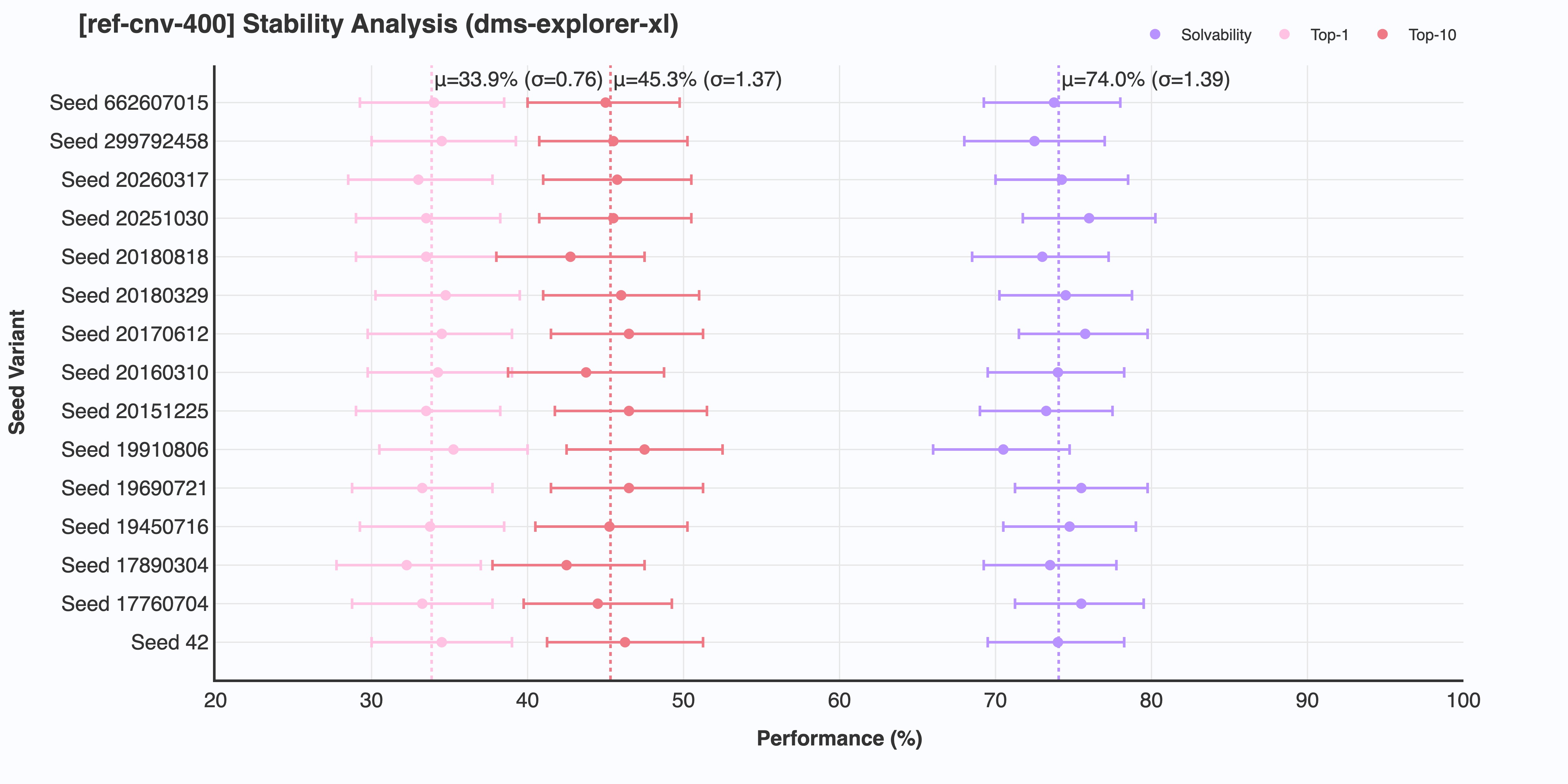}
\caption{\textbf{Benchmark stability analysis for \texttt{ref-cnv-400}.} As with other benchmarks, we evaluated 15 candidate subsets to characterize the variance from random sampling. The selected seed (662607015) yields a benchmark with performance metrics near the central tendency of all possible samples, providing a stable and fair basis for model comparison.}
\label{fig:ref-cnv-400-seed-stability}
\end{figure}

\begin{figure}[h]
\centering
\includegraphics[width=\textwidth]{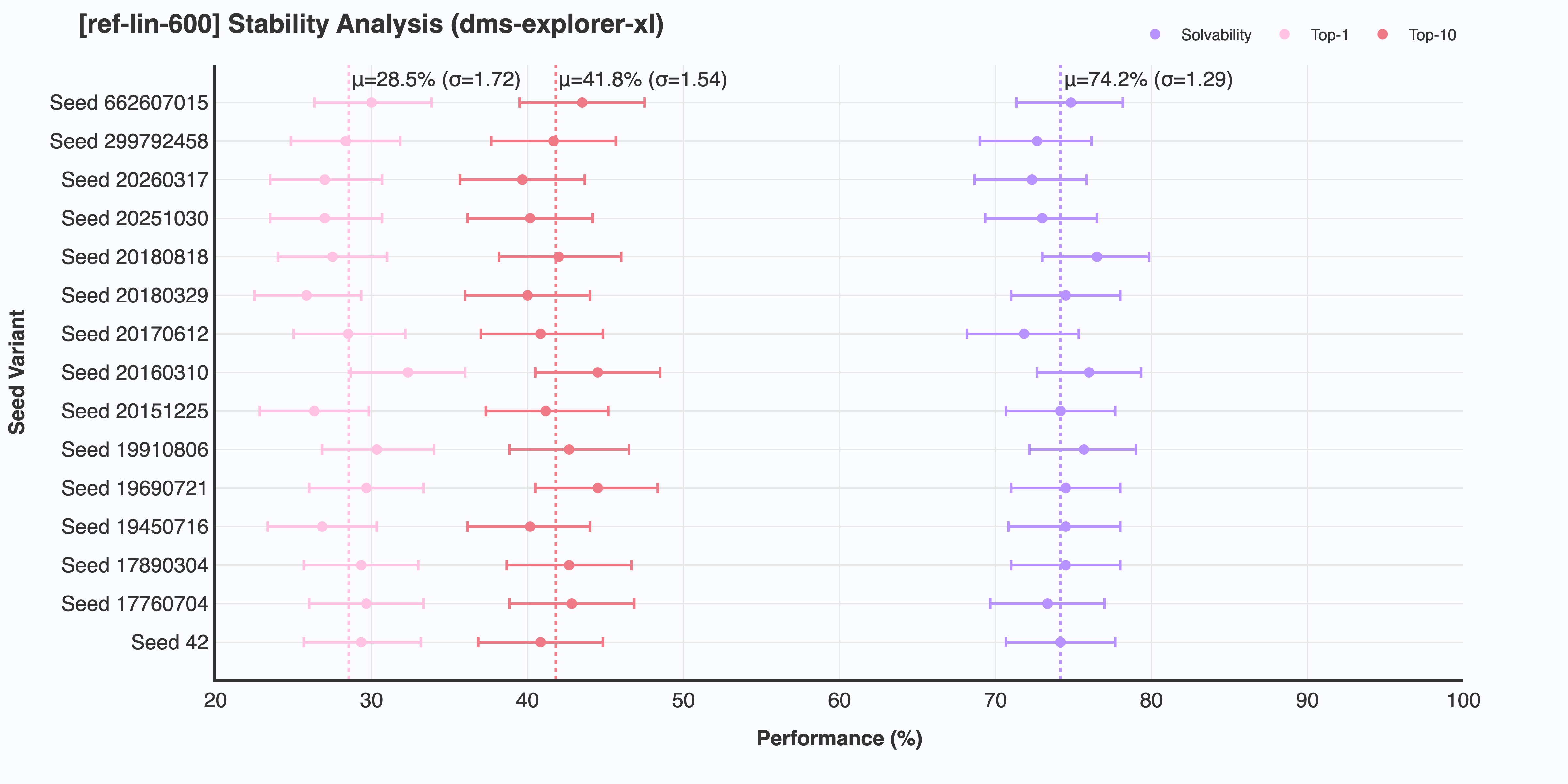}
\caption{\textbf{Benchmark stability analysis for \texttt{ref-lin-600}.} The plot demonstrates the range of performance outcomes resulting purely from the choice of random seed in benchmark sampling. Our selection of a seed (17890304) with a low Z-score deviation from the multi-seed average ensures that our reported results are generalizable and not an artifact of a fortuitous or pessimistic random sample.}
\label{fig:ref-lin-600-seed-stability}
\end{figure}

\paragraph{Supplementary Figures S2-S5}
These figures (\ref{fig:mkt-cnv-160-seed-stability}, \ref{fig:mkt-lin-500-seed-stability}, \ref{fig:ref-cnv-400-seed-stability}, \ref{fig:ref-lin-600-seed-stability}) detail the statistical stability analysis performed to select the final seed for each curated benchmark. By evaluating a reference model on 15 candidate subsets for each benchmark, we selected the seed that produced a benchmark with performance metrics closest to the multi-seed average, ensuring our results are robust against sampling artifacts.

\clearpage
\section{Supplementary Tables}
\label{sec:si-tables}

\begin{table}[h]
  \centering
  \small 
\setlength{\tabcolsep}{4pt} 
  \begin{tabular}{llcl}
  \toprule
      \textbf{Model} & \textbf{Stock Source} & \textbf{Approx. Size} & \textbf{Notes} \\
  \midrule
  MCTS\cite{mcts_2018} & Curated & 423 000 & Compounds from ZINC15 and Reaxys \\
  DFPN\cite{dfpn} & USPTO & 977 000 & All molecules from the patent training data  \\
  Retro*\cite{retrostar} & eMols & 231 000 000 & size of eMolecules screening in 2020 \\
  Self-Improved Retro\cite{selfimproved_retro_2021} & eMols & 231 000 000 & \\
  EG-MCTS\cite{egmcts_2023} & eMols & 231 000 000 & \\
  GRASP\cite{grasp_2022} & eMols & 231 000 000 & \\
  RetroGraph\cite{retrograph} & eMols & 231 000 000 & \\
  DreamRetroEr\cite{dreamretroer_2025} & eMols & 231 000 000 & \\
  SynLLaMa\cite{synllama_2025} & Enamine & 230 000 & \\
  MultiStepTTL\cite{multistepttl_2023} & Curated & 534 000 & Enamine + MolPort \\
  RetroChimera\cite{retrochimera_2025} & eMols & 23 100 000 & size of eMolecules screening in 2025 \\
  \bottomrule
  \end{tabular}
  \vspace{0.3em}
    \caption{
    \textbf{Inconsistency in Starting Material Stocks Across Major Retrosynthesis Models.} 
    The size and composition of stock sets used to define a "solved" route vary by over 1000x, ranging from curated commercial catalogs to massive, speculative screening libraries. This disparity makes direct comparison of reported solvability scores between models unreliable.
  }
    \label{tab:stock_sets}
\end{table}

\begin{table}[h]
\centering
\begin{tabular}{llcl}
  \toprule
  \textbf{Benchmark} & \textbf{Description} & \textbf{N Targets} & \textbf{Stock Set} \\
  \midrule
  \multicolumn{4}{l}{\textit{Chemist-Aligned Series (mkt-)}} \\
  mkt-lin-500 & Linear routes (len 2-6) & 500 & ASKCOS Buyables \\
  mkt-cnv-160 & Convergent routes (len 2-5) & 160 & ASKCOS Buyables \\
  \midrule
  \multicolumn{4}{l}{\textit{Developer-Aligned Series (ref-)}} \\
  ref-lin-600 & Linear routes (len 2-7) & 600 & PaRoutes n5 \\
  ref-cnv-400 & Convergent routes (len 2-5) & 400 & PaRoutes n5 \\
  ref-lng-84 & Long routes (len 8-10) & 84 & PaRoutes n1+n5 \\
  \bottomrule
\end{tabular}
\vspace{0.3em}
\caption{
  \textbf{Curated Benchmarks for Targeted Retrosynthesis Evaluation.}
  The \texttt{mkt-} series assesses practical utility using a commercial stock, while the \texttt{ref-} series enables fair algorithmic comparison using patent-derived stocks. The \texttt{lin-} and \texttt{cnv-} benchmarks contain an equal number of routes sampled from each length category, enabling a direct, unweighted analysis of performance that is not skewed by the natural prevalence of shorter routes.
  }
  \label{tab:benchmarks}
\end{table}

\begin{table}[h]
\centering
\setlength{\tabcolsep}{4pt} 
\begin{tabular}{
  l
  S[table-format=2.1] @{\,[\,} S[table-format=3.1] @{,\,} S[table-format=3.1] @{\,]}
  @{\qquad}
  S[table-format=2.1] @{\,[\,} S[table-format=2.1] @{,\,} S[table-format=2.1] @{\,]}
  @{\qquad}
  S[table-format=2.1] @{\,[\,} S[table-format=2.1] @{,\,} S[table-format=2.1] @{\,]}
}
\toprule
\textbf{Model} & \multicolumn{3}{c}{\textbf{Stock Termination (\%)}} & \multicolumn{3}{c}{\textbf{Top-1 Acc. (\%)}} & \multicolumn{3}{c}{\textbf{Top-10 Acc.
(\%)}} \\
\midrule
DMS Explorer XL & 96.3 & 93.1 & 98.8 & 21.9 & 15.6 & 28.7 & 44.4 & 36.9 & 51.9 \\
AiZynF MCTS & 98.1 & 95.6 & 100.0 & 3.1 & 0.6 & 6.3 & 8.8 & 4.4 & 13.1 \\
AiZynF MCTS (High) & 99.4 & 98.1 & 100.0 & 3.1 & 0.6 & 6.3 & 7.5 & 3.8 & 11.9 \\
AiZynF Retro* (High) & 99.4 & 98.1 & 100.0 & 0.0 & 0.0 & 0.0 & 4.4 & 1.3 & 7.5 \\
AiZynF Retro* & 98.8 & 96.9 & 100.0 & 1.3 & 0.0 & 3.1 & 4.4 & 1.3 & 8.1 \\
Retro* & 98.8 & 96.9 & 100.0 & 4.4 & 1.3 & 8.1 & 4.4 & 1.3 & 8.1 \\
Retro* (High) & 100.0 & 100.0 & 100.0 & 4.4 & 1.3 & 8.1 & 4.4 & 1.3 & 8.1 \\
Syntheseus LocalRetro & 95.6 & 92.5 & 98.8 & 0.0 & 0.0 & 0.0 & 1.3 & 0.0 & 3.1 \\
ASKCOS & 93.1 & 88.8 & 96.9 & 0.6 & 0.0 & 1.9 & 1.3 & 0.0 & 3.1 \\
\bottomrule
\end{tabular}
\vspace{0.3em}
\caption{\textbf{Baseline evaluation on \texttt{mkt-cnv-160} using a rigid single-ground-truth paradigm.} These results correspond to an evaluation where only the full, original patent route is considered correct. As discussed in the main text, this rigid definition unfairly penalizes models for finding valid shorter routes, leading to a significant underestimation of accuracy. This table serves as the baseline for the corrected results presented in main text Table 2. Brackets indicate 95\% confidence intervals. The interactive leaderboard is available on \href{https://syntharena.ischemist.com/leaderboard?benchmarkId=cmisbzwn80000zfdd10l1xe0k}{SynthArena: link}}
\label{tab:mkt-cnv-160-single-gt}
\end{table}

\begin{table}[h]
\centering
\setlength{\tabcolsep}{4pt} 
\begin{tabular}{
  l
  S[table-format=2.1] @{\,[\,} S[table-format=2.1] @{,\,} S[table-format=3.1] @{\,]}
  @{\qquad}
  S[table-format=2.1] @{\,[\,} S[table-format=2.1] @{,\,} S[table-format=2.1] @{\,]}
  @{\qquad}
  S[table-format=2.1] @{\,[\,} S[table-format=2.1] @{,\,} S[table-format=2.1] @{\,]}
}
\toprule
\textbf{Model} & \multicolumn{3}{c}{\textbf{Stock Termination (\%)}} & \multicolumn{3}{c}{\textbf{Top-1 Acc. (\%)}} & \multicolumn{3}{c}{\textbf{Top-10 Acc.
(\%)}} \\
\midrule
DMS Explorer XL & 97.2 & 95.6 & 98.6 & 27.6 & 23.6 & 31.6 & 50.2 & 45.8 & 54.6 \\
AiZynF MCTS & 97.6 & 96.2 & 98.8 & 4.2 & 2.4 & 6.0 & 14.0 & 11.0 & 17.2 \\
AiZynF MCTS (High) & 98.4 & 97.2 & 99.4 & 4.4 & 2.6 & 6.2 & 11.6 & 8.8 & 14.6 \\
AiZynF Retro* & 98.0 & 96.6 & 99.2 & 1.6 & 0.6 & 2.8 & 9.0 & 6.6 & 11.6 \\
AiZynF Retro* (High) & 99.0 & 98.0 & 99.8 & 1.2 & 0.4 & 2.2 & 7.2 & 5.0 & 9.6 \\
Retro* & 99.8 & 99.4 & 100.0 & 9.0 & 6.6 & 11.6 & 9.0 & 6.6 & 11.6 \\
Retro* (High) & 99.8 & 99.4 & 100.0 & 9.0 & 6.6 & 11.6 & 9.0 & 6.6 & 11.6 \\
Syntheseus LocalRetro & 94.0 & 91.8 & 96.0 & 1.0 & 0.2 & 2.0 & 5.0 & 3.2 & 7.0 \\
ASKCOS & 94.4 & 92.4 & 96.4 & 0.8 & 0.2 & 1.6 & 3.8 & 2.2 & 5.6 \\
\bottomrule
\end{tabular}
\vspace{0.3em}
\caption{\textbf{Baseline evaluation on \texttt{mkt-lin-500} using a rigid single-ground-truth paradigm.} Performance on the linear route benchmark using the original patent route as the sole definition of correctness. Consistent with the convergent set, these scores systematically underestimate the performance of models that identify shorter, valid pathways. This table serves as the baseline for the corrected results presented in main text Table 3. Brackets indicate 95\% confidence intervals. The interactive leaderboard is available on \href{https://syntharena.ischemist.com/leaderboard?benchmarkId=cmisbzzmb000019dd3k8yeild}{SynthArena: link}}
\label{tab:mkt-lin-500-single-gt}
\end{table}

\begin{table}[h]
\centering
\small
\setlength{\tabcolsep}{4pt} 
\begin{tabular}{
  l
  S[table-format=2.1] @{\,[\,} S[table-format=2.1] @{,\,} S[table-format=3.1] @{\,]}
  @{\quad}
  S[table-format=2.1] @{\,[\,} S[table-format=2.1] @{,\,} S[table-format=2.1] @{\,]}
  @{\quad}
  S[table-format=2.1] @{\,[\,} S[table-format=2.1] @{,\,} S[table-format=2.1] @{\,]}
  @{\quad}
  c
}
\toprule
\textbf{Model} & \multicolumn{3}{c}{\textbf{Stock Termination (\%)}} & \multicolumn{3}{c}{\textbf{Top-1 Acc. (\%)}} & \multicolumn{3}{c}{\textbf{Top-10 Acc.
(\%)}} & \textbf{Time/Target (s.)} \\
\midrule
DMS Explorer XL & 77.5 & 73.3 & 81.5 & 35.8 & 31.0 & 40.5 & 46.8 & 41.8 & 51.5 & 19.8 \\
AiZynF MCTS (High) & 87.3 & 84.0 & 90.5 & 19.0 & 15.3 & 23.0 & 32.5 & 28.0 & 37.3 & 29.6 \\
Retro* & 96.5 & 94.5 & 98.3 & 31.5 & 27.0 & 36.0 & 31.5 & 27.0 & 36.0 & 1.8 \\
Retro* (High) & 99.5 & 98.8 & 100.0 & 31.5 & 27.0 & 36.0 & 31.5 & 27.0 & 36.0 & 2.5 \\
AiZynF MCTS & 81.3 & 77.3 & 85.0 & 17.3 & 13.5 & 21.0 & 27.8 & 23.5 & 32.3 & 9.2 \\
AiZynF Retro* & 87.5 & 84.3 & 90.8 & 8.0 & 5.5 & 10.8 & 21.0 & 17.0 & 25.0 & 47.9 \\
AiZynF Retro* (High) & 94.0 & 91.5 & 96.3 & 5.8 & 3.5 & 8.0 & 15.3 & 11.8 & 18.8 & 153.7 \\
Syntheseus LocalRetro & 74.8 & 70.3 & 79.0 & 6.0 & 3.8 & 8.5 & 13.3 & 10.0 & 16.8 & 15.6 \\
\bottomrule
\end{tabular}
\vspace{0.3em}
\caption{\textbf{Controlled comparison on convergent routes confirms architectural performance signatures.} Performance on the \texttt{ref-cnv-400} benchmark. As a developer-aligned evaluation, this set uses the original PaRoutes n5 stock. The results are highly consistent with those on the chemist-aligned \texttt{mkt-cnv-160} set (Table 2), reinforcing the central observation of a performance divergence. Search-based methods excel at satisfying the stock termination condition, while the end-to-end model is more proficient at reproducing the structure of known synthetic plans. Brackets indicate 95\% confidence intervals. The full, interactive leaderboard is available on \href{https://syntharena.ischemist.com/leaderboard?benchmarkId=cmisc07hg00005iddmvjgxdau}{SynthArena: link}}
\label{tab:ref-cnv-400}
\end{table}

\begin{table}[h]
\centering
\small
\setlength{\tabcolsep}{4pt} 
\begin{tabular}{
  l
  S[table-format=2.1] @{\,[\,} S[table-format=2.1] @{,\,} S[table-format=3.1] @{\,]}
  @{\quad}
  S[table-format=2.1] @{\,[\,} S[table-format=2.1] @{,\,} S[table-format=2.1] @{\,]}
  @{\quad}
  S[table-format=2.1] @{\,[\,} S[table-format=2.1] @{,\,} S[table-format=2.1] @{\,]}
  @{\quad}
  c
}
\toprule
\textbf{Model} & \multicolumn{3}{c}{\textbf{Stock Termination (\%)}} & \multicolumn{3}{c}{\textbf{Top-1 Acc. (\%)}} & \multicolumn{3}{c}{\textbf{Top-10 Acc.
(\%)}} & \textbf{Time/Target (s.)} \\
\midrule
DMS Explorer XL & 76.8 & 73.3 & 80.2 & 30.3 & 26.7 & 34.2 & 44.8 & 40.8 & 48.8 & 18.5 \\
Retro* (High) & 97.8 & 96.7 & 99.0 & 25.8 & 22.3 & 29.3 & 25.8 & 22.3 & 29.3 & 3.5 \\
Retro* & 95.8 & 94.2 & 97.3 & 25.5 & 22.0 & 29.0 & 25.5 & 22.0 & 29.0 & 1.7 \\
AiZynF MCTS & 82.8 & 79.8 & 85.8 & 10.2 & 7.8 & 12.7 & 21.5 & 18.2 & 24.8 & 8.2 \\
AiZynF MCTS (High) & 88.8 & 86.3 & 91.3 & 9.7 & 7.3 & 12.2 & 21.5 & 18.3 & 24.8 & 25.1 \\
AiZynF Retro* & 88.2 & 85.5 & 90.7 & 7.3 & 5.3 & 9.5 & 21.7 & 18.3 & 25.0 & 29.4 \\
AiZynF Retro* (High) & 95.0 & 93.2 & 96.7 & 6.2 & 4.3 & 8.2 & 16.5 & 13.7 & 19.5 & 109.9 \\
Syntheseus LocalRetro & 67.2 & 63.3 & 71.0 & 5.0 & 3.3 & 6.8 & 13.0 & 10.3 & 15.7 & 15.4 \\
\bottomrule
\end{tabular}
\vspace{0.3em}
\caption{\textbf{Controlled comparison on linear routes confirms the divergence between stock termination and accuracy.} Performance on the \texttt{ref-lin-600} benchmark. This developer-aligned set uses the original PaRoutes n5 stock, isolating algorithmic performance from the choice of commercial stock. The results corroborate the findings from the main text (Table 3): search-based models achieve high stock termination rate, while the sequence-based DirectMultiStep model exhibits substantially higher Top-K accuracy. This demonstrates that the observed divergence is a robust architectural signature, not an artifact of a specific stock set. Brackets indicate 95\% confidence intervals. The full, interactive leaderboard is available on \href{https://syntharena.ischemist.com/leaderboard?benchmarkId=cmisc09u000007qddhq6eczgb}{SynthArena: link}}
\label{tab:ref-lin-600}
\end{table}

\begin{table}[h]
\centering
\small
\setlength{\tabcolsep}{4pt} 
\begin{tabular}{
  l
  S[table-format=2.1] @{\,[\,} S[table-format=2.1] @{,\,} S[table-format=3.1] @{\,]}
  @{\quad}
  S[table-format=2.1] @{\,[\,} S[table-format=2.1] @{,\,} S[table-format=2.1] @{\,]}
  @{\quad}
  S[table-format=2.1] @{\,[\,} S[table-format=2.1] @{,\,} S[table-format=2.1] @{\,]}
  @{\quad}
  c
}
\toprule
\textbf{Model} & \multicolumn{3}{c}{\textbf{Stock Termination (\%)}} & \multicolumn{3}{c}{\textbf{Top-1 Acc. (\%)}} & \multicolumn{3}{c}{\textbf{Top-10 Acc.
(\%)}} & \textbf{Time/Target (s.)}\\
\midrule
DMS Explorer XL & 72.6 & 63.1 & 82.1 & 45.2 & 34.5 & 56.0 & 50.0 & 39.3 & 60.7 & 31.3 \\
Retro* (High) & 94.0 & 89.3 & 98.8 & 8.3 & 2.4 & 14.3 & 8.3 & 2.4 & 14.3 & 8.6 \\
Retro* & 86.9 & 79.8 & 94.0 & 8.3 & 2.4 & 14.3 & 8.3 & 2.4 & 14.3 & 3.9 \\
AiZynF Retro* (High) & 76.2 & 66.7 & 84.5 & 0.0 & 0.0 & 0.0 & 0.0 & 0.0 & 0.0 & 134.2 \\
AiZynF Retro* & 61.9 & 51.2 & 72.6 & 0.0 & 0.0 & 0.0 & 0.0 & 0.0 & 0.0 & 31.7 \\
AiZynF MCTS (High) & 59.5 & 48.8 & 70.2 & 0.0 & 0.0 & 0.0 & 0.0 & 0.0 & 0.0 & 27.2 \\
AiZynF MCTS & 46.4 & 35.7 & 57.1 & 0.0 & 0.0 & 0.0 & 0.0 & 0.0 & 0.0 & 8.9 \\
Syntheseus LocalRetro & 36.9 & 27.4 & 47.6 & 0.0 & 0.0 & 0.0 & 0.0 & 0.0 & 0.0 & 16.6 \\
\bottomrule
\end{tabular}
\vspace{0.3em}
\caption{\textbf{Stress test on long routes provides stark evidence of the "complexity cliff."} Performance on the \texttt{ref-lng-84} benchmark, a challenging set composed of all routes of length 8-10 from the PaRoutes evaluation sets. This benchmark is designed to probe model performance at the limits of planning complexity. The results provide the clearest evidence of the architectural trade-offs discussed in the main text: the Top-K accuracy of all search-based models collapses to near-zero. In contrast, the sequence-based DirectMultiStep model retains substantial accuracy, demonstrating its robustness on long-range planning tasks. Brackets indicate 95\% confidence intervals. The full, interactive leaderboard is available on \href{https://syntharena.ischemist.com/leaderboard?benchmarkId=cmisc03z600003oddt8xnrkvq}{SynthArena: link}}
\label{tab:ref-lng-84}
\end{table}

\clearpage
\section{Supplementary Figures}
\label{sec:si-figures}

\end{document}